\documentclass[10pt,twocolumn,letterpaper]{article}

\usepackage{iccv}
\usepackage{times}
\usepackage{epsfig}
\usepackage{graphicx}
\usepackage{amsmath}
\usepackage{amssymb}

\usepackage[pagebackref=true,breaklinks=true,letterpaper=true,colorlinks,bookmarks=false]{hyperref}

\label{sub:subsection_name}

\iccvfinalcopy % *** Uncomment this line for the final submission

\newcommand{\norm}[1]{\left\lVert #1 \right\rVert}
\ificcvfinal\fi
\DeclareMathOperator*{\argmax}{arg\,max}

\def\lc{\left\lceil}   
\def\rc{\right\rceil}
\begin{document}

%%%%%%%%% TITLE
\title{An End-to-End Approach to Natural Language Object Retrieval \\via Context-Aware Deep Reinforcement Learning}
\author{Fan Wu$^\S$\thanks{This work was done while F. Wu was visiting University of Technology Sydney.}\hspace{2em}Zhongwen Xu$^\dag$\hspace{2em}Yi Yang$^\dag$\\
$^\S$Zhejiang University\hspace{2em}$^\dag$University of Technology Sydney\\
	{\tt\small \{jxwufan,zhongwen.s.xu,yee.i.yang\}@gmail.com}
}
\maketitle
%\thispagestyle{empty}

%%%%%%%%% ABSTRACT
\begin{abstract}
We propose an end-to-end approach to the natural language object retrieval task, which localizes an object within an image according to a  natural language description, \ie, referring expression. Previous works divide this problem into two independent stages: first, compute region proposals from the image without the exploration of the language description; second, score the object proposals with regard to the referring expression and choose the top-ranked proposals. The object proposals are generated independently from the referring expression, which makes the proposal generation redundant and even irrelevant to the referred object. In this work, we train an agent with deep reinforcement learning, which learns to move and reshape a bounding box to localize the object according to the referring expression. 
We incorporate both the spatial and temporal context information into the training procedure. 
By simultaneously exploiting local visual information, the spatial and temporal context and the referring language a priori, the agent selects an appropriate action to take at each time. 
A special action is defined to indicate when the agent finds the referred object,  and terminate the procedure. 
 We evaluate our model on various datasets, and our algorithm significantly outperforms the compared algorithms. 
 Notably, the accuracy improvement of our method over the recent method GroundeR and SCRC on the ReferItGame dataset are $7.67\%$ and $18.25\%$, respectively.   
\end{abstract}

%%%%%%%%% BODY TEXT

\section{Introduction}
Convolutional Neural Networks (ConvNets) has shown phenomenal results  \cite{krizhevsky2012imagenet,simonyan2014very,szegedy2015going,he2016deep} for many computer vision applications. With ConvNets, the object detection tasks have been practiced in a more accurate model better than ever. Existing detection algorithms aim to detect a predefined object category from the given image. As a result, 
detection based retrieval systems usually take the target object name as the query, which largely ignore the context information within an image.
 In the real world, however, the rich information that a user is searching could be more than what a single object name can describe. Compared to object names,  language description contains context information such as the relative location of the object, \eg, ``the book on your \textit{left-hand side}'', or a specific part of an object, \eg, ``\textit{his face}''. With language descriptions, one can even specify detailed attributes of the object of interest, \eg ``the man in \textit{middle jeans }and \textit{T-shirt}''. Therefore, natural language provides users with more powerful tools than the scheme of adopting object name as the query.

\begin{figure}[t]
\begin{center}
\includegraphics[width=\linewidth]{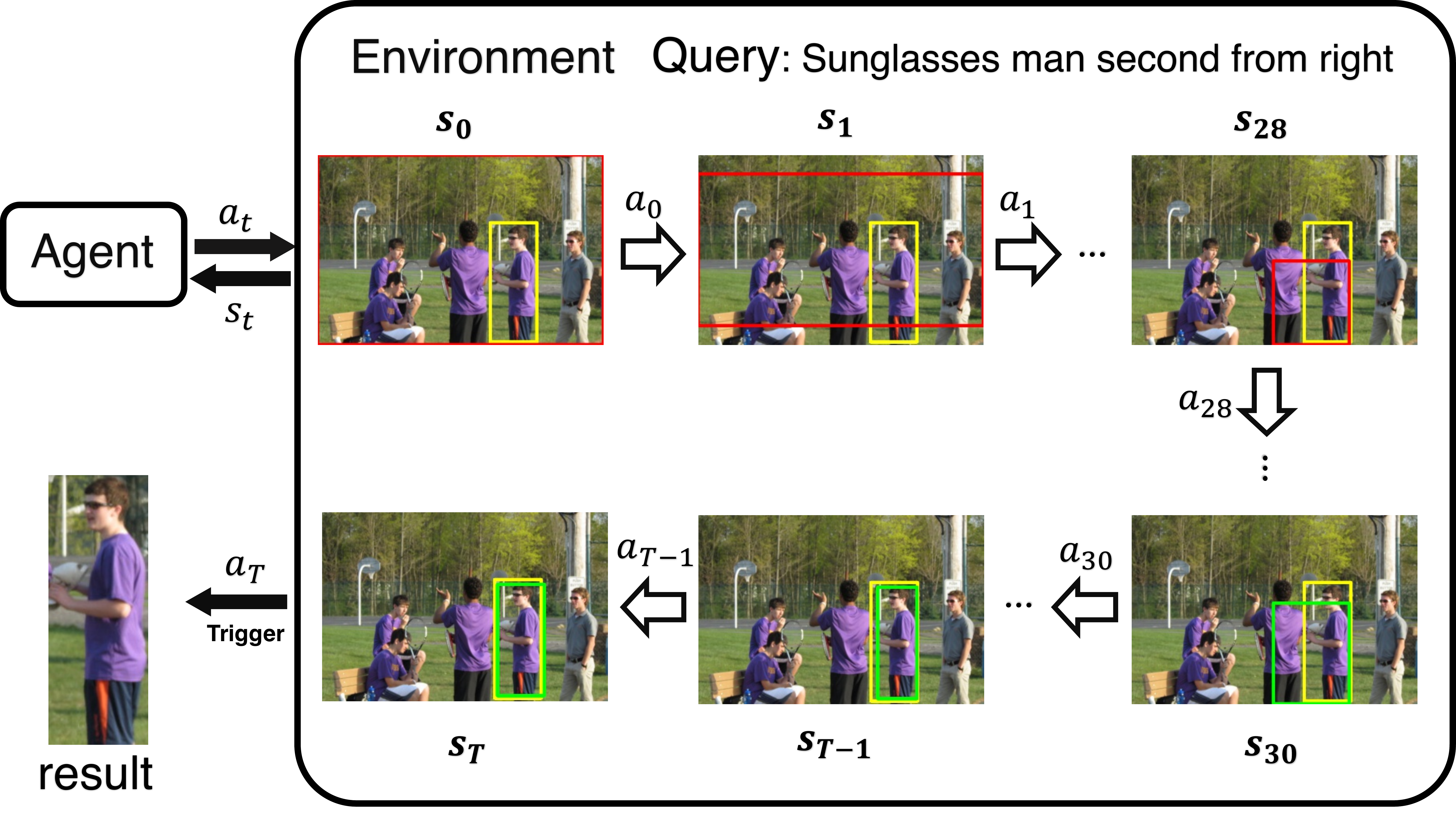}
\end{center}
   \caption{Illustration of the proposed context aware reinforcement learning framework. The yellow box is the ground truth. The bounding box generated by the agent at each time step is green if the Intersection-over-Union (IoU) value with the ground truth box is greater than 0.5, while red otherwise. Best viewed in color.}
\label{fig:overview}
\end{figure}

In this work, we propose a new method of natural language object retrieval. The goal is to localize a referred object in an image or a set of images according to a language description, which can be interpreted as a new type of cross-media retrieval~\cite{yang2009ranking}.
A typical and straightforward way is to divide the task into two non-overlapping phases. In the first phase, a set of object region proposals are generated as which has been done in~\cite{girshick2014rich,girshick2015fast,ren2015faster}. If the algorithm uses handcrafted features, \eg, EdgeBoxes~\cite{ZitnickECCV14edgeBoxes}, the quality of proposals may not be good. \cite{yu2016joint} and \cite{yu2016modeling} use the Fast R-CNN~\cite{girshick2015fast} and SSD~\cite{liu2016ssd} to generate object detection results as object proposals. However, this type of approaches rely heavily on the training data of object proposals and are restricted to the predefined object categories. As a result, these algorithms~\cite{yu2016joint,yu2016modeling} can only deal with the predefined objects and are not extendable to natural language queries containing new objects and complex reasoning of relative location. 
In the second phase, these methods adopt a ranking function to locate the region which best matches the description. 
The limitation of this kind of method is that the two critical phases are conducted independently. In this case, the training process is not well aligned, leading to suboptimal solution for the retrieval task. Furthermore, those approaches usually rely on a large number of proposals to guarantee a satisfactory recall for the target object, which drastically increases redundancies and degrades the discriminative performance of the ranking function.

Inspired by the recent successes of deep reinforcement learning \cite{mnih2015human,mnih2016asynchronous,silver2016mastering},  we propose to train a neural network for natural language object retrieval in an end-to-end manner.
As illustrated in Figure~\ref{fig:overview}, our method adopts a top-down approach to localize the referred object. Specifically, we define different actions for an agent to change the shape and location of a bounding box. 
The ``agent''  takes one of those predefined actions according to the spatial and temporal context, the local image feature as well as the natural language a priori at each time step, until an optimal result is reached, \ie, the agent takes a special action (denoted to as a ``trigger'') and stops the process. It is worthwhile highlighting the following aspects of the proposed method. 

First, our approach performs natural language object retrieval in an end-to-end manner without the pre-computation of proposals, which could be very noisy and redundant.  Different from our approach, existing natural object retrieval methods \cite{hu2016natural,rohrbach2016grounding,yu2016modeling} either use handcrafted features or ConvNet features to generate proposals in the first phase. The performance of handcrafted features is comparatively poor. On the other hand, the ConvNet based detectors can only deal with a limited number of predefined object categories. Our end-to-end approach exploits language information and visual information in a joint framework, thereby being able to leverage the mutual benefits of the two inputs for training. Moreover, our approach also avoids the non-trivial task of tuning the number of proposals. Instead, the network decides to stop searching the object by selecting the ``trigger'' action, thus it constructs a dynamic length search procedure per query.

Second, our approach generates a series of ``experiences'' to better use the training information under the deep reinforcement learning paradigm~\cite{mnih2015human,mnih2016asynchronous}. Image-level context information is complementary to local information within a bounding box~\cite{hu2016natural}.
This context is presumably important in natural language object retrieval, especially when the language description contains relative locations. Therefore, we propose to use image-level ConvNet representation as \textit{spatial context}, and explicitly encode such information into the ``experience''. 
Further, a Recurrent Neural Network (RNN) is added into the policy and value networks to track the \textit{temporal context}, \ie, the history states that the agent has encountered. This temporal context would help the agent avoid entering similar mistakes in the previous time steps. 
The existing approaches~\cite{hu2016natural,rohrbach2016grounding} merely use the labeled images for training. The ``context-aware experience" in our algorithm is generated at each time step after the ``agent'' takes action.  The number of ``context-aware experience" is greater than the number of labeled images, and have more diversified information. Meanwhile, as shown in Figure~\ref{fig:overview}, the difference between the bounding boxes at state $s_T$ and state $s_{T-1}$ is subtle but has different IoU values with the ground truth. These subtle differences are also encoded in the ``experience''. In that way, our method is able to exploit subtle changes of bounding boxes for a better result. 

Third, environment state, agent action and reward function are three key factors for reinforcement learning~\cite{sutton1998reinforcement}. Different from a typical deep reinforcement learning scenario, \eg, game playing, computer vision tasks have no well-defined reward function provided by the environment. To address this issue, we define a simple yet effective reward function for the agent. In addition, a potential based reward strategy is also adopted to improve the training speed. Besides, the visual content of an environment are quite similar in game playing scenario. For example, An Atari game~\cite{mnih2015human} has less diversified visual information. Our task is very different because the environment states presented to the agent keep changing dramatically, \ie, natural language queries and images can be very different from one to another. We take advantage of this diversity nature by paralleling a series of agents and environments when collecting experiences in training as practiced in \cite{mnih2016asynchronous}.

\section{Related Work}
\noindent{\bf Object Detection.} Using object proposals to detect object inside image has been validated to be an effective approach. Girshick~\etal~\cite{girshick2014rich} propose R-CNN framework to crop and warp the region proposals generated from off-the-shelf object proposal algorithms, then score each region based on its ConvNet feature. Girshick~\cite{girshick2015fast} introduces Fast R-CNN, especially the ``RoI pooling'' technique to share the feature computation among all the proposal regions, which enhances the processing speed of object detection significantly. Ren~\etal~\cite{ren2015faster} further improve the object detection system by replacing the external proposal algorithm with a ConvNet which applies sliding windows on the feature maps and outputs bounding boxes. All the methods described above are limited to predefined categories, which cannot immediately generalize to other categories.

\noindent{\bf Deep Reinforcement Learning.} Recently, deep reinforcement learning has many breakthroughs. Mnih~\etal~\cite{mnih2015human} utilize deep neural networks, \ie, Deep Q-learning Network (DQN), to parametrize an action-value function to play Atari games, reaching human-level performance. Silver~\etal~\cite{silver2016mastering} use policy network and value network to play Go and beat a world-class professional player. Mnih~\etal~\cite{mnih2016asynchronous} tackle the training efficiency issue of deep reinforcement learning with an asynchronous approach, making it feasible to train strong agents in a short time on a single machine with CPU only. On the aspect of computer vision applications, Caicedo~\etal~\cite{caicedo2015active} and Jie~\etal~\cite{jie2016tree} apply the DQN proposed in \cite{mnih2015human} to generate object proposals in an image with an MDP setting similar to ours. Yeung~\etal~\cite{yeung2016end} apply a policy gradient method called REINFORCE~\cite{williams1992simple} to detect actions in videos.

\noindent{\bf Vision and Language.}
Recurrent neural networks have been widely used in vision and language tasks, starting from image captioning tasks~\cite{vinyals2015show}. Recently, Johnson~\etal~\cite{johnson2016densecap} propose a model which could be trained end-to-end to localize objects and produce description for dense regions, namely the dense captioning task. Mao~\etal~\cite{mao2016generation} propose a discriminative training strategy to generate unambiguous descriptions for objects. Yu~\etal~\cite{yu2016modeling, yu2016joint} further improve the result on the dense captioning task. Hu~\etal~\cite{hu2016natural}, Nagaraja~\etal\cite{nagaraja16refexp} and  Rohrbachcite~\etal~\cite{rohrbach2016grounding} focus on retrieving an object inside an image given a language description referring the object. Rohrbach~\etal~\cite{rohrbach2016grounding} use an attention model to localize the language description in an image by choosing the region that could be best used to reconstruct the description. Different from our end-to-end approach, previous works on the natural language object retrieval task \eg~\cite{hu2016natural,rohrbach2016grounding}, use an algorithm to generate object proposals or use the result of an object detection algorithm directly.  The referring expression is not utilized in the object proposal or detection procedure.

%------------------------------------------------------------------------

\section{Context-Aware Deep Reinforcement Learning}
In this section, we detail the proposed context-aware deep reinforcement learning algorithm. 
\subsection{Markov Decision Process (MDP)}
\begin{figure}[t]
\begin{center}
\includegraphics[width=0.75\linewidth]{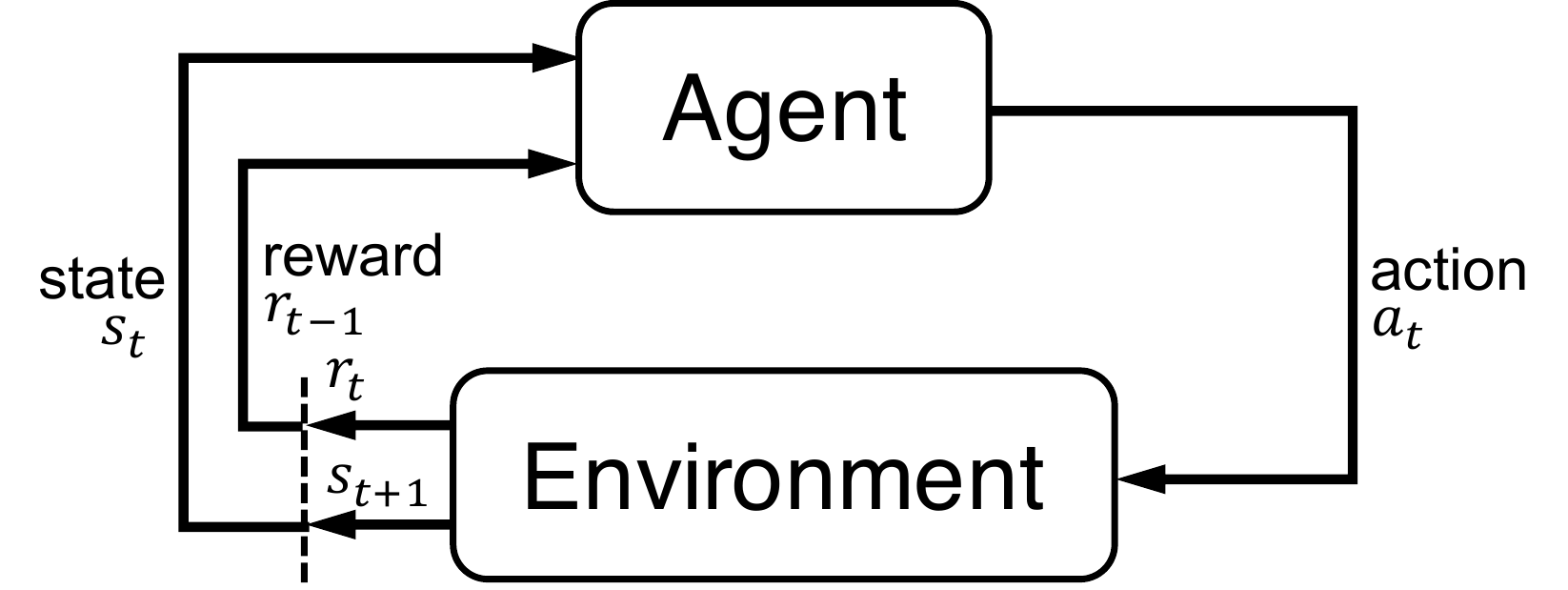}
\end{center}
   \caption{An illustration of the interaction between an agent and an environment~\cite{sutton1998reinforcement}.}
\label{fig:agent_env}
\end{figure}

A Markov Decision Process (MDP) is a sequential decision process, which describes how an agent could interact with an environment and what will happen after each interaction. We denote the MDP as $(S, A, R, \gamma)$, where $S$ is a set of states of the environment, $A$ is a set of actions of which the agent could choose from to act on the environment, $R :S \times A \rightarrow \mathbb{R}$ is a reward function that maps a state-action pair $(s, a)$ to a reward $r \in \mathbb{R}$, and $\gamma \in (0,1]$ is a discount factor determining the decay rate in calculating the cumulative discounted reward of the entire trajectory.

The agent interacts with the environment on time step $0, 1, \cdots, T$, where $T$ corresponds to termination. From these interactions, a trajectory $\{(s_t, a_t, r_{t})\}_{t=0, \cdots, T}$ is generated. At each time step $t\in [0,T]$, the agent takes an action $a_{t} \in A$ based on the current state $s_t$ of the environment. After receiving the action $a_t$ from the agent, the environment transits from state $s_t$ to state $s_{t+1}$. And the agent receives a reward $r_{t}$ from the environment. The agent is to maximize the expected cumulative discounted reward $\mathbb{E}[R_t]$ for each of the state $s_t$, where 

\begin{align}
R_t=\sum_{k=0}^{T-t}{\gamma^k r_{k + t}}.\label{eqn:rt}
\end{align}

In our case, 
the agent changes the size and the position of a bounding box inside the image by using a set of actions to localize an object according to the referring expression. The details are given as follows.
\begin{figure}[t]
\begin{center}
\includegraphics[width=\linewidth]{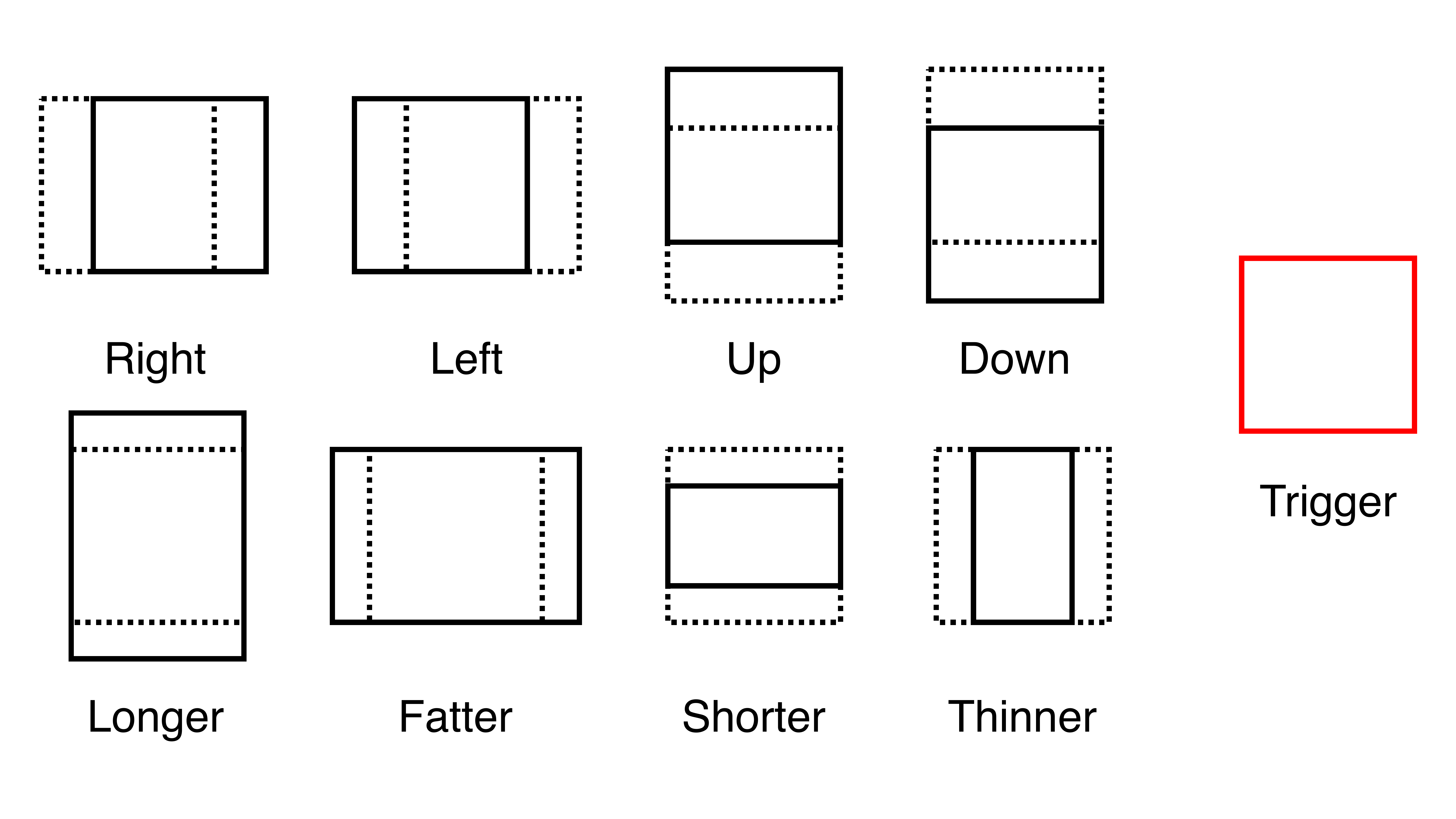}
\end{center}
   \caption{The actions for the agent. The dashed line indicates the bounding box before the action. The solid line is the bounding box after the action. The trigger indicates termination.}
\label{fig:action}
\end{figure}

\begin{figure*}[t]
\begin{center}
\includegraphics[width=0.9\linewidth]{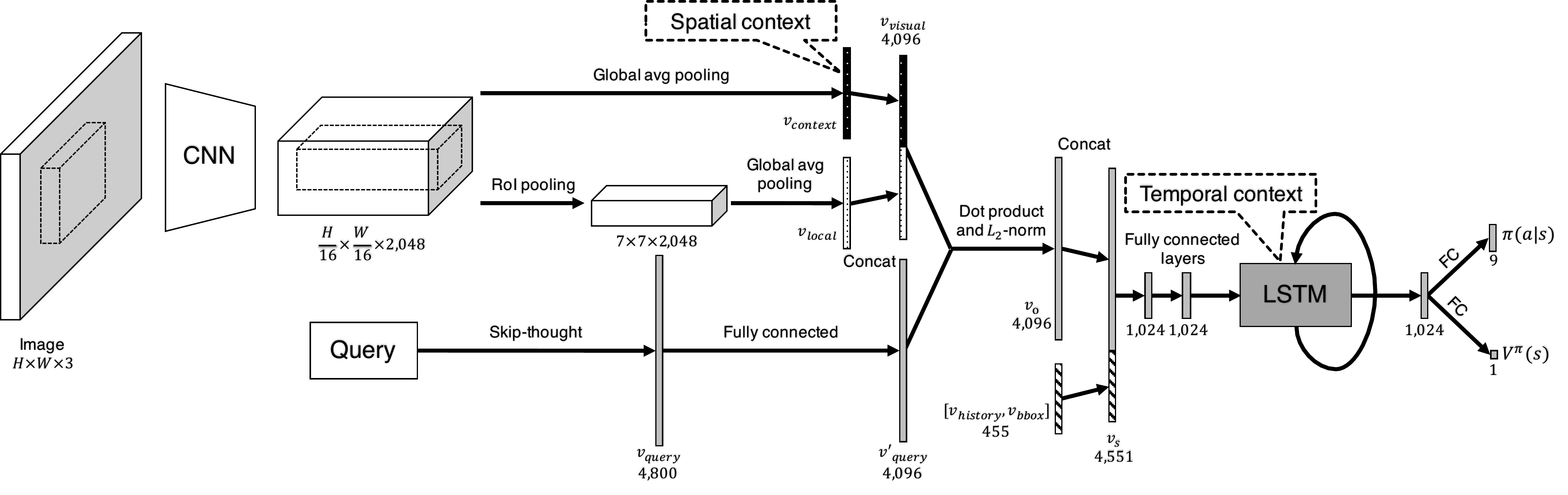}

\end{center}
   \caption{The proposed context-aware policy and value network. The spatial context is computed by applying global average pooling to the entire feature maps. The temporal context is encoded in the state of the LSTM. The outputs of the network are the policy $\pi(a|s)$ and the value $V^{\pi}(s)$. FC indicates the Fully Connected layer. The numbers under each vector indicate the dimensions of the vector.}
\label{fig:policy}
\end{figure*}

\subsection{Environment States and Actions}
We define the environment state $s=\{I, Q, \text{bbox}\}$  as a combination of the image $I$, the referring expression $Q$ and the bounding box $\text{bbox}=[x_{0}, y_{0}, x_{1}, y_{1}]$ for localizing the target, where $(x_{0},y_{0})$ and $(x_{1},y_{1})$ are the top-left and bottom-right coordinates respectively. The bounding box is initialized to cover the whole image.

The nine actions as shown in Figure~\ref{fig:action} can be categorized into three groups. Four actions of the first group move the location of the bounding box. Another four actions of the second group change the shape of the bounding box. We have an additional action ``trigger'' to indicate that the agent has achieved an optimal result. Each action moves the top-left point and bottom-right point of the bounding box to adjust it, where the change is proportional to the height and the width of current bounding box. We denote the absolute changes at $x$ and $y$ coordinate directions as $|\Delta x| = \delta \cdot W_{\text{bbox}}$ and $|\Delta y| = \delta \cdot H_{\text{bbox}}$, where $W_{\text{bbox}}$ and $H_{\text{bbox}}$ are the width and the height of the bounding box respectively. For the movement actions, we set $\delta = 0.2$. For the shape changing actions in the second row, we set the factor $\delta = 0.1$.
For example, if the agent takes an ``UP'' action, the bounding box will be changed from $[x_{0}, y_{0}, x_1, y_1]$ to $[x_0, y_{0}- 0.2 \cdot(y_1-y_0), x_1, y_1- 0.2 \cdot (y_1-y_0)]$.

\subsection{Reward Shaping}
In computer vision applications of reinforcement learning, we need to define our reward function, instead of using the reward signals provided by the environment directly. An appropriate reward function for natural language object retrieval task is an essential factor of the success of this work.

For a state-action pair $(s, a)$, we define our reward function $R(s_t, a_t)$ as follows:

\begin{align}
   R(s_t,a_t) &= \Bigg \{ \begin{tabular}{lll}
    $R'(s_t,a_t)+F(s_t,a_t)$ &$\text{if}~~a_t \neq \text{trigger}$ \\  
    \\
    $E(s_t,a_t)$ &  \text{if}~~$a_t = \text{trigger}$ 
  \end{tabular}.
\end{align}
In the above formulation, $R'(s_t,a_t)$, $F(s_t,a_t)$ and the $E(s_t,a_t)$ are defined as follows,
\begin{align}
R'(s_t,a_t) &= \Bigg \{ \begin{tabular}{lll}
    $\text{IoU}(s_{t+1})$ &$\text{if}~~\text{IoU}(s_{t+1}) > \text{IoU}(s)$ \\ 
    &$\forall s \in s_{0\dots t}$ \\
    $-p$ &$\text{otherwise}$ 
  \end{tabular} \\
   E(s_t,a_t) &= \Bigg \{ \begin{tabular}{ll}
    $\eta$ &$\text{if}~~\text{IoU}(s_t) > \tau$ \\  
    \\
    $-\eta$ &$\text{otherwise}$  \\
  \end{tabular} 
\end{align}
\begin{align}
F(s_t,a_t)&=-\Phi(s_t) + \gamma \Phi(s_{t+1}) \label{eqn:F}\\
\Phi(s_t)&= \text{IoU}(s_t)\label{eqn:phi},
\end{align}
In the equations above,  $s_{t+1}$ is the state of environment after the agent takes action $a_t$, $\eta$ is the quantity of the reward for ``trigger'', $\tau$ is a threshold of IoU value, and $-p$ is the penalty imposed on the agent when it makes no progress. The $\text{IoU}$ function measures the Intersection-over-Union between the current bounding box and the ground truth box of the target in the current state.

The basic reward $R'(a_t,s_t)$ equals to IoU$(s_{t+1})$ when the new state $s_{t+1}$ has a higher IoU value than all the other states the agent has encountered so far. Otherwise, a penalty $-p$ will be given to the agent. We use $p=0.05$. Intuitively, this reward function encourages the agent move towards high IoU value states. However, this reward signal is rarely positive. It is hard for the agent to find the goal only with this reward. We add an additional reward, called potential based reward $\Phi(s)$. It is constructed from $\text{IoU}(s)$ function as shown in Eqn~(\ref{eqn:F}) and Eqn~(\ref{eqn:phi}). This kind of reward can accelerate the training process~\cite{ng1999policy}. Lastly, the termination reward function $E$ is decided by the IoU value in the termination state $s_T$. If IoU$(s_T)>\tau$, a positive reward $\eta$ will be generated. Otherwise, the agent will receive a penalty $-\eta$. We set $\tau=0.5$ and $\eta=1.0$ empirically. Our discount factor $\gamma$ is set to $0.99$ as in most deep reinforcement learning literatures~\cite{mnih2015human,mnih2016asynchronous}. 

\begin{figure*}[t]
\begin{center}
\includegraphics[width=\linewidth]{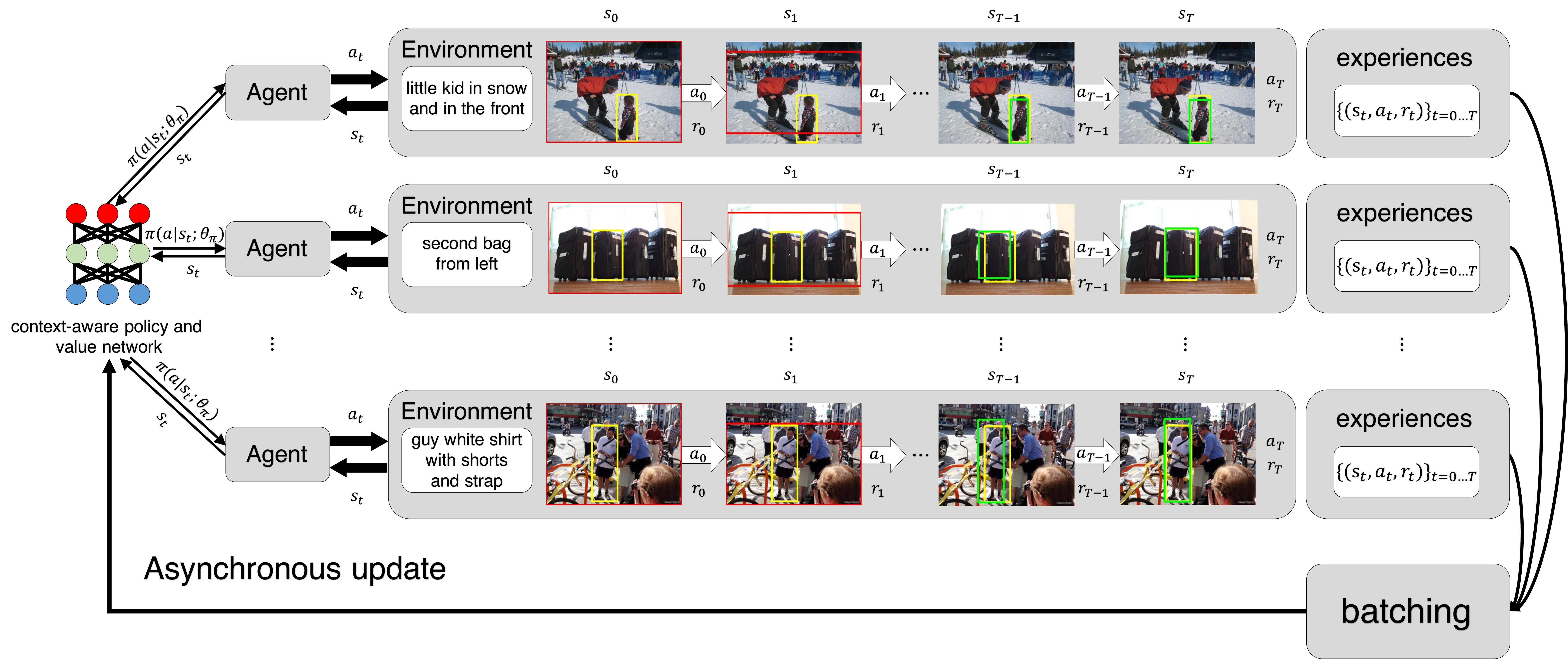}

\end{center}
   \caption{Overview of our training pipeline, we use multiple agents with environments to inference on the current network in parallel. An agent with each query generates a sequence of experiences. Note that the termination time $T$ for each query is variable. A data collector collects the training tuples from all agents, batches the data to update the context-aware policy and value network  shown in Figure~\ref{fig:policy} asynchronously. The color of the bounding box is green if its IoU between the ground truth box is over $0.5$, and red otherwise. Best viewed in color.}
\label{fig:train}
\end{figure*}

\subsection{Policy and Value Networks}
Our agent uses a policy function $\pi(a|s)$ to get a distribution of actions given a state $s$, and then decides which action to take according to the probabilities over actions. The agent also uses a value function $V^{\pi}(s) = \mathbb{E}\left[R_t|s_t=s\right]$ to estimate the expected cumulative discounted reward $R_t$ from any state $s$ under the policy $\pi$. As Figure~\ref{fig:policy} shows, we use a neural network to parametrize the policy function and value function. These two functions share a common network until the last fully-connected (FC) layer~\cite{jaderberg2016reinforcement, mnih2016asynchronous}. The network takes the state of the environment as input, and outputs the distribution $\pi(a|s)$ over discrete actions and the value estimation $V^{\pi}(s)$ of the state $s$. The ReLU activations are applied between the FC layers.

Our network uses the ResNet-152~\cite{he2016deep} which is pre-trained on the ImageNet dataset~\cite{ILSVRC15} to extract the visual feature. To encode the spatial context information, we feed the image with width $W$ and height $H$ into a modified ResNet-152 model which has been applied with the atrous algorithm~\cite{chen14semantic} on the $\text{conv}_5$ stage, resulting in image feature maps of size $\frac{H}{16} \times \frac{W}{16} \times 2048$. The feature maps are then fed to a RoI pooling layer~\cite{girshick2015fast,jie2016tree} to compute the local feature maps inside the bounding box of size $7\times 7 \times 2048$. We feed these two feature map groups to two global average pooling layers~\cite{he2016deep} to obtain two visual feature vectors $v_\text{context}$ and $v_\text{local}$. $v_\text{context}$ is the spatial context, and it is only computed only once for all time steps. We denote $v_\text{visual}=[v_\text{context}, v_\text{local}]$.
For the language aspect, we utilize skip-thought vectors~\cite{kiros2015skip} trained on the BookCorpus dataset~\cite{zhu2015aligning} to encode the query description. We denote the encoded query feature as $v_\text{query}$, which is then projected to $v_\text{query}' \in \mathbb{R}^{4,096}$ by a FC layer. After applying dot product and $L_2$-norm to $v_\text{query}'$ and $v_\text{visual}$, we obtain the observation of the current state as $v_{o}=\frac{v_\text{query}'\cdot v_\text{visual}}{\norm{v_\text{query}'\cdot v_\text{visual}}}$.

However, after the operations above, the computed vector $v_o$ may lose considerable amount of information which is originally in the state $s$. Thus we propose to leverage the temporal context which tracks the states that the agent has encountered as well as all the actions that the agent has taken. In this paper, 50 previous actions are recorded, which  generates a history vector $v_\text{history} \in \mathbb{R}^{450}$. Following \cite{mao2016generation}, we define $v_\text{bbox}=[\frac{x_{0}}{W},\frac{y_{0}}{H},\frac{x_{1}}{W},\frac{y_{1}}{H},\frac{S_\text{bbox}}{S_\text{image}}]$, where $S_\text{bbox}$ and $S_\text{image}$ are the areas of bounding box and image. We use $v_\text{s}=[v_{o}, v_\text{history}, v_\text{bbox}]$ as the vector representation of state. After passing the $v_\text{s}$ to two FC layers with the same output size of 1,024, a Long Short-Term Memory (LSTM)~\cite{hochreiter1997long} cell 
with Layer Normalization~\cite{ba2016layer} is used to track the past states~\cite{mnih2016asynchronous,jaderberg2016reinforcement,hausknecht2015deep}. The state inside the LSTM cell is the temporal context for subsequent decision making. Specifically, the output of LSTM will be passed to two FC layers without activation function respectively. One FC layer outputs the policy $\pi(a|s)$ (followed by the softmax operation). The other FC layer outputs the value $V^\pi(s)$.

\subsection{Training}
An on-policy algorithm~\cite{sutton1998reinforcement} interacts with the environment, then uses its own experiences $\{(s, a, r)\}$ to update the current policy. Using a single agent to collect experiences from the environment may get data highly correlated.  Updating from such experiences would lead the agent to a suboptimal solution. Therefore, we adopt the asynchronous advantage actor-critic (A3C) method~\cite{mnih2016asynchronous} which uses multiple agents associated with environments to collect data in parallel and updates the policy asynchronously.

As Figure~\ref{fig:train} shows, we use multiple agents that share a common and global neural network. We denote the policy function and value function from the network as $\pi(a | s;\theta_{\pi})$ and $V(s ;\theta_{v})$, where $\theta_{\pi}$ is the parameters of the network outputting the policy function, and $\theta_{v}$ are the parameters of the network outputting value function. For one query, an agent uses the current network to interact with the environment constructed by the query. The agent generates an episode $\{(s_t, a_t, r_t)\}_{t=0\dots T}$ for training.  After a query is processed by an agent, the agent will randomly select another query to process.
The network parameters are asynchronously updated. The actions in an episode may be chosen by different parameters.

Every $N$ consecutive experiences in every episode are grouped. At the time step $t$, each $(s_t, a_t, r_t)$ is converted to a training tuple $(s_t,a_t,R_t')$, where $R_t'$ is defined as: 

\begin{align}
R_t'=\Bigg \{ \begin{tabular}{ll}
    $\sum_{k=t}^{t_{m}(t)-1} \gamma^{k-t} r_{k} + \gamma^{t_m(t)-t}V(s_{t_{m}(t)})$\\ 
    $~~~~~~~~~~~~~~~~~~~~~~~~~~~~~~~~~~~~~~~~~~~~\text{if}\ t+N\leq T$ \\
    \\
     $\sum_{k=t}^{T} \gamma^{k-t} r_{k}$ $~~~~~~~~~~~~~~~~~~~~~\text{otherwise}$ \label{eqn:Rt}
  \end{tabular}\end{align}
In Eqn~(\ref{eqn:Rt}), $t_{m}(t)=\lc \frac{t}{N}\rc \cdot N$. We set $N=5$ as in~\cite{mnih2016asynchronous}. All the tuples are collected in parallel, and used to optimize in batch mode as follows:

\begin{align}
\begin{split}
\theta_\pi\leftarrow \theta_\pi +\alpha&((R_t'-V(s_t;\theta_v))\nabla_{\theta_\pi}\log\pi(a_t|s_t;\theta_\pi)\\
&+\beta\nabla_{\theta_\pi}H(\pi(\cdot|s_t;\theta_\pi))),
\end{split} \\
\theta_v\leftarrow \theta_v - \alpha&\nabla_{\pi_v}(R_t-V(s_t;\theta_v))^2,
\end{align}
%http://tex.stackexchange.com/questions/56627/multiple-alignment
where $\alpha$ is the learning rate, $H(\pi(\cdot|s_t;\theta_\pi))$ is the entropy of the policy~\cite{mnih2016asynchronous}, $\beta$ is a hyper parameter, $(R_t'-V(s_t;\theta_v))\nabla_{\theta_\pi}\log\pi(a_t|s_t;\theta_\pi)$ is policy gradient~\cite{williams1992simple} which gives the direction to update the policy such that the agent gets more rewards.

The network is trained by the ADAM optimizer~\cite{kingma2014adam}. We set $\alpha = 10^{-4}$ and $\beta=10^{-2}$ during the training. The learning rate $\alpha$ is halved once.

\section{Experiments}

\subsection{Running Environment Details}
We implement our model with TensorFlow~\cite{abadi2016tensorflow} and tensorpack\footnote{\url{https://github.com/ppwwyyxx/tensorpack}}, running on one NVIDIA GTX-1080 GPU. At training time, we use 50 processes to run agents with environments and one process to run policy and value network. The agent processes communicate with the network process via IPC interface provided by operating system. We make our code and the trained models publicly available
upon acceptance.
\subsection{Preprocessing and Testing}
We preprocess every image, ground truth box and query text in all dataset. For the images and ground truth boxes, we followed \cite{ren2015faster} to resize every image such that its shorter side length equals 600 pixels, and the ground truth boxes are resized proportionally. We filter out non-alphanumeric characters in the query and convert the rest of the characters to their lowercase as in common practices.

At test time, we fix the policy and value network, and use a single agent to process each query. To get the result deterministically, given a state $s_t$ of the current environment, the agent uses the network to get the probabilities $\pi(a|s_t;\theta_{\pi})$ over actions, then takes the action $a_t=\argmax_a \pi(a |s_t;\theta_{\pi})$ which has the highest probability. The agent stops taking actions at the time step $T$ when it uses the ``trigger''. The bounding box $\text{bbox}$ inside the $s_T$ is the result of the algorithm for the query.

\begin{figure}
\begin{center}
\includegraphics[width=\linewidth]{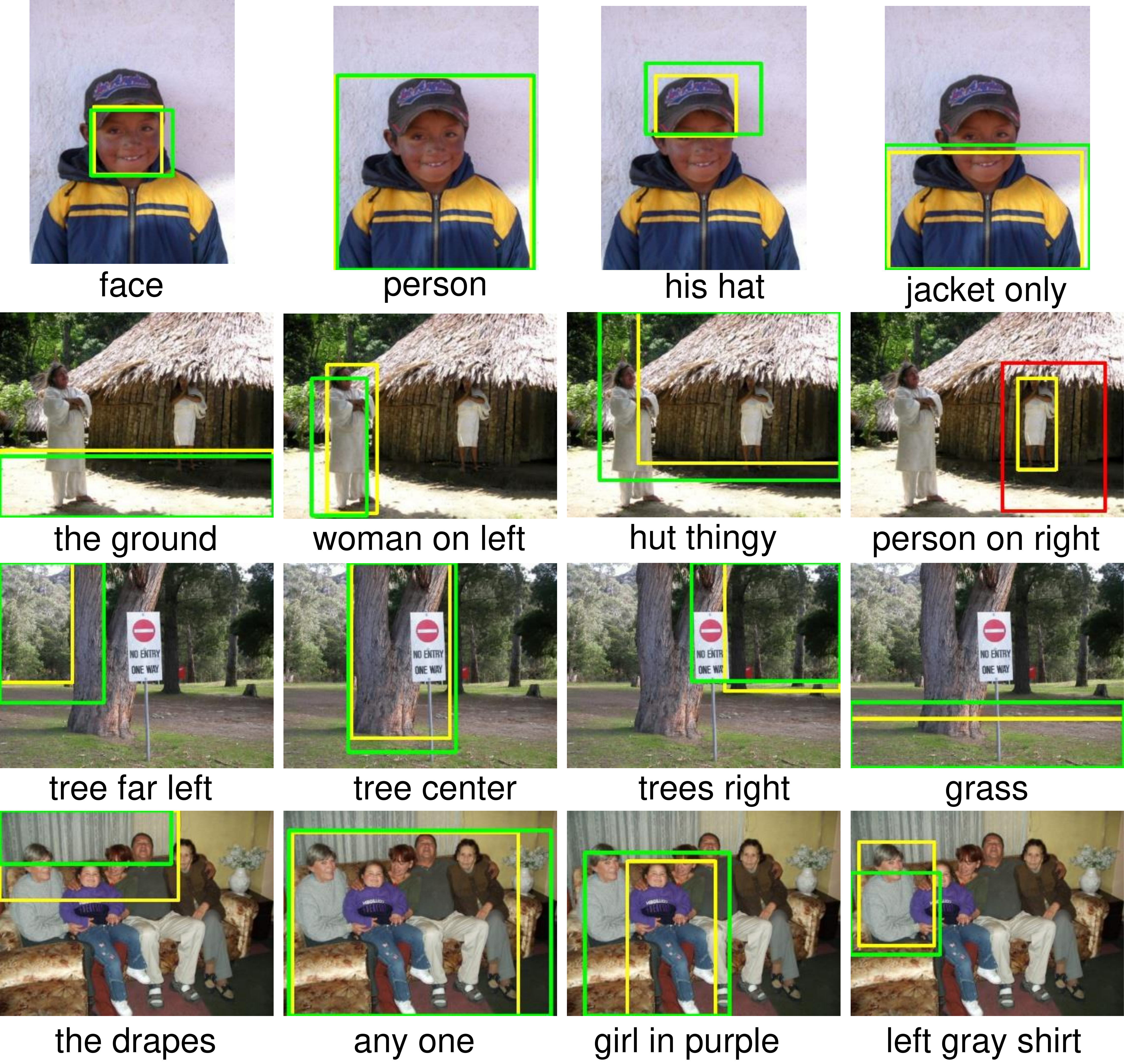}
\end{center}
   \caption{Examples from the testing set of the ReferItGame dataset. The yellow box is the ground truth. The color of the bounding box is green if the result is correct, and red otherwise. Best viewed in color.}
\label{fig:kid}
\end{figure}
\subsection{Experiments on the ReferItGame Dataset}
Following \cite{hu2016natural} and \cite{rohrbach2016grounding}, we first test our framework on the ReferItGame dataset~\cite{kazemzadeh2014referitgame}. The ReferItGame dataset contains 20,000 images from ImageCLEF IAPR image retrieval dataset.  For each object, the dataset uses a segmentation region to describe its shape and location information. In total, there are 238 object categories in the dataset. 
Since the objects in the original dataset are localized with pixel-level segmentations instead of bounding boxes, Hu~\etal~\cite{hu2016natural} converted the segmentation data of each object to a bounding box, then split the whole dataset to two subsets the trainval set and testing set. We use the meta-data and split provided by~\cite{hu2016natural}. The processed dataset contains 59,976 instances in the trainval set and 60,105 in the testing set. We train our model on the trainval set.
During testing, we use the trained agent to give a bounding box for each query in the testing set.

\begin{table}
\begin{center}
\begin{tabular}{|l|c|}
\hline
Method & Accuracy \\
\hline
LRCN~\cite{donahue2015long} & 8.59\% \\
CAFFE-7K~\cite{guadarrama2014open}& 10.38\%\\
SCRC~\cite{hu2016natural} & 17.93\%\\
GroundeR~\cite{rohrbach2016grounding} & 28.51\%\\
\hline
%Ours w/o LSTM global & 36.03\%\\
%Ours w/o global & 35.59\%\\
Ours & \textbf{36.18\%}\\
\hline
\end{tabular}
\end{center}
\caption{Accuracy on the ReferItGame dataset.}\label{Tab:referit}
\end{table}
As shown in Table~\ref{Tab:referit}, our method outperforms all previous approaches. This result proves our end-to-end model could better exploit the connection between the visual data and language a priori.

Figure \ref{fig:kid} shows some examples of our result on the testing set of the ReferItGame dataset. Given an image with different natural language queries, our algorithm can correctly locate these queried objects even if the object is a part of another object, \eg, a person's hat. In addition, when the query is about multiple objects or very complex, our algorithm still achieves good performance,  \eg, when the query is ``any one''.

%------------------------------------------------------------------------

\subsection{Experiments on The RefCOCO, RefCOCO+ and Google Refexp (RefCOCOg) Datasets}
\begin{figure}
\begin{center}
\includegraphics[width=\linewidth]{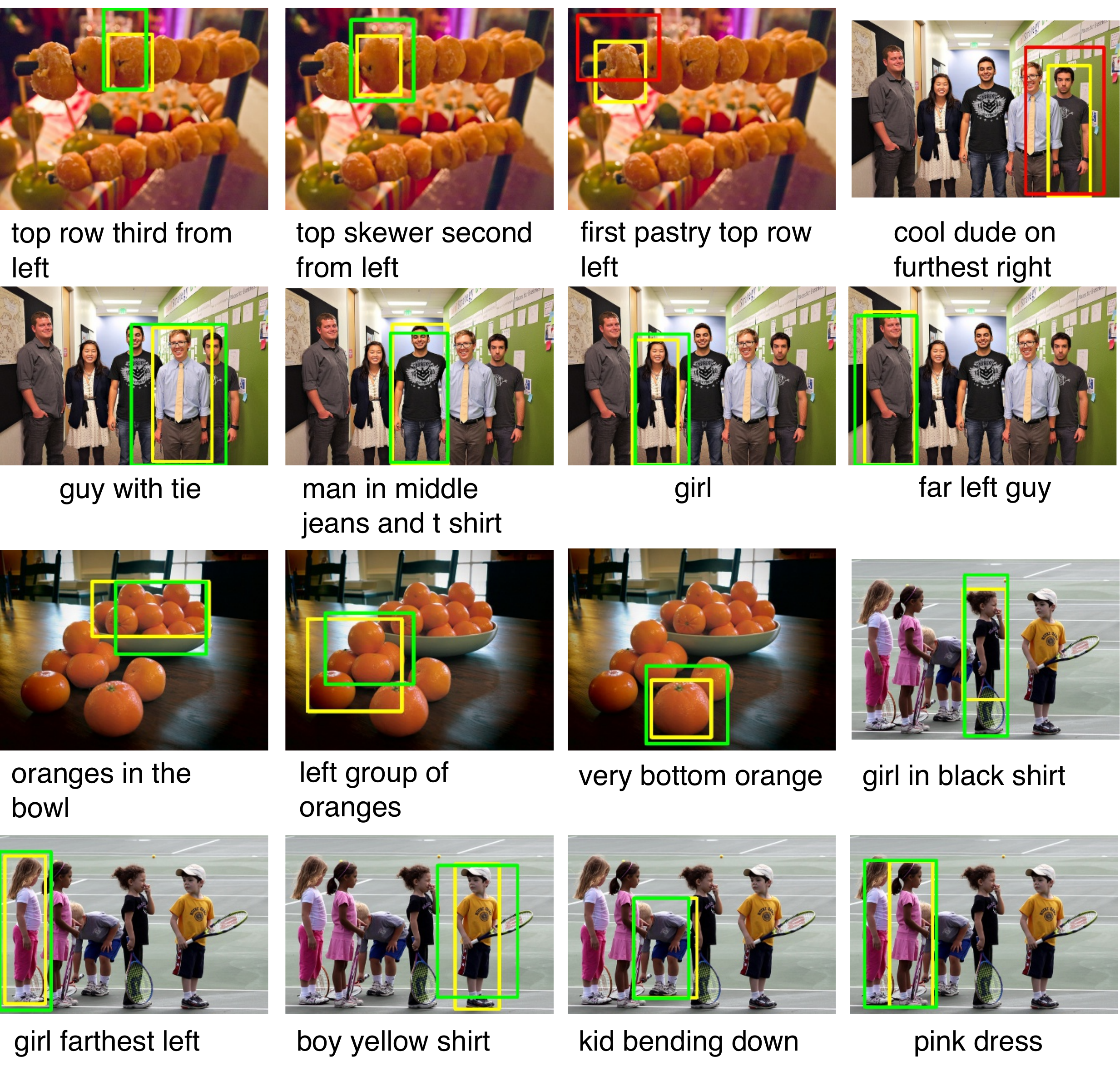}
\end{center}
   \caption{Examples from the testing set of the RefCOCO dataset. The yellow box is the ground truth. The color of the bounding box is green if the result is correct, and red otherwise. Best viewed in color.}
\label{fig:refcocoexm}
\end{figure}

\begin{table}[h]
\scriptsize
\begin{center}
\begin{tabular}{| l | c | c | c |}
\hline
\multicolumn{4}{|c|}{RefCOCO} \\
\hline
Method &\ \ Test A\ \  &\ \ Test B\ \ &\ \ Validation\ \  \\
\hline
SCRC with ~~10 Proposals ~\cite{hu2016natural}     & 14.58\%   & 18.39\%   & 16.39\%  \\
%\hline
SCRC with ~~50 Proposals ~\cite{hu2016natural}     & 19.36\%   & 20.92\%   & 19.39\%  \\
%\hline
SCRC with 100 Proposals ~\cite{hu2016natural}     & 18.47\%   & 20.16\%   & 19.02\%  \\
%\hline
SCRC with 200 Proposals ~\cite{hu2016natural}     & 16.46\%   & 18.37\%   & 16.88\%  \\
\hline
Ours   & \textbf{54.78}\%   & \textbf{41.58}\%     & \textbf{48.19}\%  \\
\hline
\end{tabular}
\end{center}
\caption{Accuracy on the RefCOCO dataset. We only used the training set of the RefCOCO dataset for training.}
\label{table:RefCOCO}
\end{table}
\begin{table}[h]
\scriptsize
\begin{center}
\begin{tabular}{| l | c | c | c |}
\hline
 \multicolumn{4}{|c|}{RefCOCO+} \\
\hline
Method&\ \ Test A\ \  &\ \ Test B\ \ &\ \ Validation\ \  \\
\hline
SCRC with ~~10 Proposals ~\cite{hu2016natural}      & 12.57\%   & 14.69\%   & 14.46\%  \\
%\hline
SCRC with ~~50 Proposals ~\cite{hu2016natural}      & 15.51\%   & 14.09\%   & 14.00\%  \\
%\hline
SCRC with 100 Proposals ~\cite{hu2016natural}     & 14.43\%   & 13.25\%   & 13.72\%  \\
%\hline
SCRC with 200 Proposals ~\cite{hu2016natural}    & 13.60\%   & 12.13\%   & 12.46\%  \\
\hline
Ours    & \textbf{40.39}\%   & \textbf{22.81}\%     & \textbf{31.93}\%  \\
\hline
\end{tabular}
\end{center}
\caption{Accuracy on the RefCOCO+ dataset. We only used the training set of the RefCOCO+ dataset for training.}
\label{table:RefCOCOp}
\end{table}

\begin{table}[h]
\scriptsize
\begin{center}
\begin{tabular}{| l | c |}
\hline
\multicolumn{2}{|c|}{RefCOCOg} \\
\hline
Method & Validation\\
\hline
SCRC with ~~10 Proposals \cite{hu2016natural}      & 15.09\%  \\
SCRC with ~~50 Proposals \cite{hu2016natural}    & 16.91\%  \\
SCRC with 100 Proposals \cite{hu2016natural}      & 16.91\%  \\
SCRC with 200 Proposals \cite{hu2016natural}     & 15.29 \%  \\
\hline
Ours   & \textbf{29.04}\%   \\
\hline
\end{tabular}
\end{center}
\caption{Accuracy on the RefCOCOg dataset. We only used the training set of the RefCOCOg dataset for training.}
\label{table:RefCOCOg}
\end{table}

\begin{table*}[h]
\scriptsize
\begin{center}
\begin{tabular}{| l | c | c | c | c | c | c | c | c |}
\hline
&  \multicolumn{3}{c|}{RefCOCO} & \multicolumn{3}{c|}{RefCOCO+} & \multicolumn{1}{c|}{RefCOCOg}\\
\cline{2-8}
&  \ ~~~Test A~~~\ \  &\ \ ~~~Test B~~~\ \ &\ \ Validation\ \  &\ \ ~~~Test A~~~\ \  &\ ~~~Test B~~~\ \  &\ Validation\ \ &\ \ Validation\ \  \\
\hline
Ours w/o spatial and temporal context   & 46.89\%   & 35.51\%     & 41.25\%   & 32.01\% & 15.91\% & 25.57 \%& 21.75\%\\
Ours w/o spatial context  & 49.78\%   & 37.33\%     & 42.90   & 36.74\% & 19.55\% & 29.11\%& 29.14\%\\
Ours full  & 54.78\%   & 41.58\%     & 48.19\%   & 40.39\% & 22.81\% & 31.93\% & 29.04\% \\
\hline
\end{tabular}
\end{center}
\caption{A comparison of the results with and without the context information.}
\label{table:ablation}
\end{table*}

\begin{table}[h]
\scriptsize
\begin{center}
\begin{tabular}{| l | c | c |}
\hline
\multicolumn{3}{|c|}{RefCOCO} \\
\hline
Method &\ \ Test A\ \  &\ \ Test B\ \ \\
\hline
SCRC with ~~10 Proposals ~\cite{hu2016natural}   & 16.14\%   & 18.96\%     \\
%\hline
SCRC with ~~50 Proposals ~\cite{hu2016natural}     & 20.93\%   & 21.22\%    \\
%\hline
SCRC with 100 Proposals ~\cite{hu2016natural} & 19.85\%& 20.41\%  \\
%\hline
SCRC with 200 Proposals ~\cite{hu2016natural}    &18.46 \%   & 18.65\%    \\
\hline
Ours   & \textbf{59.66}\%   & \textbf{44.49}\%  \\

\hline
\end{tabular}
\end{center}
\caption{Accuracy on the RefCOCO dataset with more training data. In this experiment, we increase the number of training data by using the combination of the training and validation sets of the RefCOCO, RefCOCO+ and RefCOCOg datasets.}
\label{table:RefCOCO_M}
\end{table}

\begin{table}[h]
\scriptsize
\begin{center}
\begin{tabular}{| l | c | c |}
\hline
\multicolumn{3}{|c|}{RefCOCO+} \\
\hline
Method&\ \ Test A\ \  &\ \ Test B\ \  \\
\hline
SCRC with ~~10 Proposals ~\cite{hu2016natural} &13.59\% & 14.69\% \\
%\hline
SCRC with ~~50 Proposals ~\cite{hu2016natural} &17.11\% & 14.77\% \\
%\hline
SCRC with 100 Proposals ~\cite{hu2016natural} &16.29\% & 14.26\% \\
%\hline
SCRC with 200 Proposals ~\cite{hu2016natural} &14.86\% & 12.44\% \\
\hline
Ours    & \textbf{47.05}\%   & \textbf{29.09}\%  \\
\hline
\end{tabular}
\end{center}
\caption{Accuracy on the RefCOCO+ dataset with more training data. In this experiment, we increase the number of training data by using the combination of the training and validation sets of the RefCOCO+, RefCOCO+ and RefCOCOg datasets.}
\label{table:RefCOCOp_M}
\end{table}
We validate our model on the RefCOCO dataset, the RefCOCO+ dataset~\cite{yu2016modeling} and the Google Refexp Dataset (RefCOCOg)~\cite{mao2016generation} in this section. It is worth noting that the referring expressions in the RefCOCO+ dataset contain no location word. 
%To make the description more general, there is no location word in the referring expressions in the RefCOCO+ dataset. %the player is not allowed to use location words when describing objects in the RefCOCO+ dataset. 
%The RefCOCOg dataset is collected by a similar scheme. %Rather than online game play, they use one worker from a group to give a description of a specified object and ask another worker from another group to check its validity in an offline setting. 
In total, the RefCOCO dataset contains 19,994 images with 142,209 descriptions for 50,000 objects. The RefCOCO+ dataset contains 19,992 images with 141,564 descriptions for 49,856 objects. The RefCOCOg dataset contains 26,711 images with 85,474 descriptions for 54,822 objects. We use the original split provided by each dataset. The RefCOCO and RefCOCO+ datasets split their testing set to two set TestA and TestB. The images in the TestA set contain multiple people, and the images in the TestB set only contain non-human objects.

We use the SCRC algorithm~\cite{hu2016natural}, which does not require extra labeled data for proposal detector training, as our baseline. In addition to the training set used in our algorithm and \cite{hu2016natural}, the authors of \cite{yu2016modeling} and \cite{yu2016joint} used a large amount of extra training data, \ie, the validation set and trainval set of MSCOCO~\cite{lin2014microsoft},  to pretrain object detectors. Therefore, we did not include the results of \cite{yu2016modeling} and \cite{yu2016joint} for a fair comparison.

We train our model using the training set of each dataset, and test our model on the testing set and validation set of those three datasets respectively. 
Specifically, both our algorithm and \cite{hu2016natural} use the training set of the RefCOCO, the RefCOCO+ and the RefCOCOg as training data. 
We report the results of the SCRC model using Top-10, Top-50, Top-100 and Top-200 proposals on the testing and validation set of the three datasets. 
The results are reported in Table~\ref{table:RefCOCO}, Table~\ref{table:RefCOCOp} and Table~\ref{table:RefCOCOg}.  We can see that our algorithm outperforms SCRC~\cite{hu2016natural} for all settings dramatically.

Figure~\ref{fig:refcocoexm} shows some of our sample outputs from the testing set of the RefCOCO dataset. Results show our algorithm could process queries contain relationships with another object as in the skewer example, queries contain multiple objects as in the orange example, and queries contain complex attributes as in the person example.

\subsection{Ablation Study of Context Information}
In this subsection we test the effects of the context, \ie, the spatial context and the temporal context, in the reinforcement learning. 
Recall that our algorithm uses an LSTM as temporal context for state tracking, and uses the image level ConvNets representation as spatial context.  We train two modified versions of our algorithm. The first one does not contain spatial and temporal context.  The other model only removes spatial context (image level ConvNets representation) from our method. We denote the model without all context infomration as ``Ours w/o spatial and temporal context''. The model with only temporal context %only without global feature input 
is denoted as ``ours w/o spatial context''.
As Table~\ref{table:ablation} shows, the ablation reveals both spatial and temporal context plays an important role in the context-aware policy and value network.

\subsection{Performance with More Training Data}
Taking the RefCOCO series dataset as an example, we show the performance improvement when the number of training data increases. To obtain more training data, we merge the training sets and the validation sets of the RefCOCO dataset, the RefCOCO+ dataset and the RefCOCOg dataset, and name it as the RefCOCOmg trainval dataset means merged dataset. Note that the testing set of the RefCOCOg dataset has not been released. %does not release the testing set.
We use the testing set of the RefCOCO+ dataset and the RefCOCO dataset as the testing data. If an image is in both the testing set and the RefCOCOmg trainval set, we will remove it from the trainval set. In total, the RefCOCOmg trainval set contains 29,456 images with 352,511 descriptions. The experiment results are shown in Table~\ref{table:RefCOCO_M} and Table~\ref{table:RefCOCOp_M}. Compared to Table~\ref{table:RefCOCO} and Table~\ref{table:RefCOCOp}, We observe that as we have more training data, the performances of our method and SCRC~\cite{hu2016natural} both increase. %both algorithm increases.
Nevertheless, our method still dramatically outperforms SCRC~\cite{hu2016natural}. Also, our method benefits more from more training data.

\section{Conclusion}

In this paper, we present an end-to-end deep reinforcement learning model for the natural language object retrieval task. Unlike previous approaches, our model leverages the context information and exploits the visual information and language a priori in a joint framework. Extensive experiments on various dataset demonstrate effectiveness of our model. Since our method does not constrain the query object in predefined categories, our method has great potential to be generalized in real world scenarios.

{\small
\bibliographystyle{ieee}
\bibliography{egbib}
}

\end{document}